\documentclass[10pt,twocolumn,letterpaper]{article}

\usepackage[pagenumbers]{cvpr} 

\usepackage{bm}
\usepackage{multirow}

\definecolor{cvprblue}{rgb}{0.21,0.49,0.74}
\usepackage[pagebackref,breaklinks,colorlinks,allcolors=cvprblue]{hyperref}

\def\paperID{9} 
\def\confName{CVPR}
\def\confYear{2025}

\title{GaussianVideo: Efficient Video Representation and Compression by Gaussian Splatting}

\author{Inseo Lee \quad Youngyoon Choi \quad Joonseok Lee\textsuperscript{*}\\
Seoul National University\\
{\tt\small \{ian.lee, youngyoon911, joonseok\}@snu.ac.kr}
}

\begin{document}

\twocolumn[{
\maketitle
\begin{center}
    \captionsetup{type=figure}
    \includegraphics[width=\textwidth]{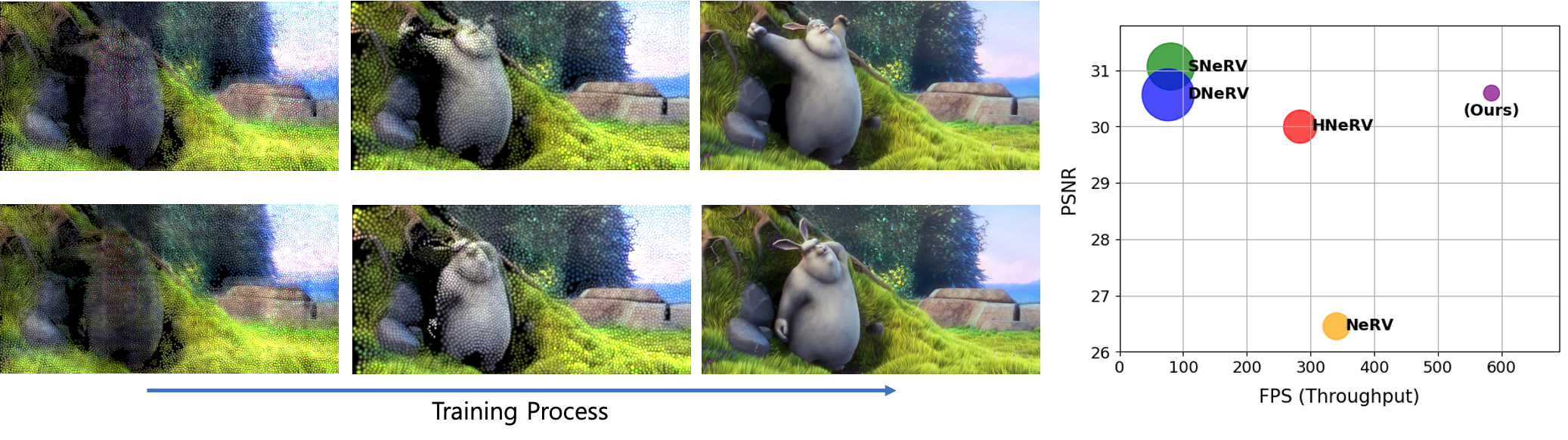}
    \captionof{figure}{(Left) Demonstration of the proposed GaussianVideo method on the Bunny Dataset, learning video representation using Gaussian Splatting.
    (Right) Experimental comparison with several state-of-the-art NeRV methods, demonstrating superior rendering speed (FPS) and competitive reconstruction quality (PSNR) of our method, with the least training time (represented as the circle size).}
\end{center}
}]

\renewcommand{\thefootnote}{}  
\footnotetext{\textsuperscript{*}Corresponding author}
\renewcommand{\thefootnote}{\arabic{footnote}}  

\begin{abstract}
Implicit Neural Representation for Videos (NeRV) has introduced a novel paradigm for video representation and compression, outperforming traditional codecs. As model size grows, however, slow encoding and decoding speed and high memory consumption hinder its application in practice. To address these limitations, we propose a new video representation and compression method based on 2D Gaussian Splatting to efficiently handle video data. Our proposed deformable 2D Gaussian Splatting dynamically adapts the transformation of 2D Gaussians at each frame, significantly reducing memory cost. Equipped with a multi-plane-based spatiotemporal encoder and a lightweight decoder, it predicts changes in color, coordinates, and shape of initialized Gaussians, given the time step. By leveraging temporal gradients, our model effectively captures temporal redundancy at negligible cost, significantly enhancing video representation efficiency. Our method reduces GPU memory usage by up to 78.4\%, and significantly expedites video processing, achieving 5.5x faster training and 12.5x faster decoding compared to the state-of-the-art NeRV methods.
\end{abstract}
\section{Introduction}
\label{sec:intro}
The increasing demand for video processing, driven by the rise of various video-sharing platforms, has intensified research efforts in this field.
Particularly, the large data volumes inherent to video content have also led to concurrent advancements in video compression techniques.
Over the past few decades, traditional codecs \cite{sullivan2012overview, wiegand2003overview} have been highly effective for video compression.
However, as the size and resolution of videos have increased, these hand-crafted codecs face challenges in scalability.

Recently, neural representation techniques have shown strong advantages for effectively representing visual content, shifting the trends in image and video processing.
By encoding a video within a neural network, Neural Representation for Video (NeRV) \cite{chen2021nerv} models have enhanced both video quality and compression capabilities.
Despite its advantages, however, relying on neural networks brings certain limitations as well.
Encoding and decoding processes tend to be slow, and the computational demands increase sharply as the model size gets larger, significantly degrading their efficiency.

To address the limitations of implicit models in the 3D domain \cite{martin2021nerf}, 3D Gaussian Splatting \cite{kerbl20233d} has been introduced, offering both fast rendering speed and high-quality results.
However, since the technique was originally developed for novel view synthesis, its application in video remains limited to Novel View Synthesis for Dynamic Scenes,
with relatively little research extending it into areas like video representation and compression.

In this paper, we aim to apply Gaussian Splatting to video representation and compression to enable fast training and rendering.
While it may initially appear trivial 
to apply 3D Gaussians directly to video data, the approach is more complex than it seems.
Unlike conventional 3D spaces defined by \(x\), \(y\), and \(z\) axes, video is represented in an \(x\), \(y\), \(t\) space, evolving over time.
Thus, capturing this dynamic nature with a fixed set of 3D Gaussians significantly reduces modeling flexibility, which, in turn, limits the overall representational capacity of the approach.
Another alternative is applying the 2D Gaussian approach used for images \cite{zhang2025gaussianimage} across all frames independently, but this approach would increase the complexity proportional to the length of video.

To address these limitations, we propose a deformable 2D Gaussian Splatting framework.
Unlike conventional 3D Gaussians, our approach leverages time-conditioned attributes-such as 2D mean, color, rotation, and scale—of Gaussians, adapting them at each time step $t$.
This enables the model to more accurately and flexibly represent the changes occurring within video frames, providing a powerful alternative to the fixed, static 3D Gaussians typically used in spatial data modeling.
Our framework efficiently capitalizes on the inherent spatial redundancy found in video content, thereby reducing memory requirements and computational overhead.

In addition, we adopt a multi-plane based \cite{cao2023hexplane, fridovich2023kplanes} encoder-decoder structure to replace the traditional MLP to enhance scalability.
This approach allows more efficient representation of high-dimensional video data, where the multi-plane structure divides the spatial and temporal components across multiple planes, ensuring that each is appropriately captured without the high parameter demands of a full multilayered perceptron (MLP).

Lastly, to further optimize our framework for dynamic video scenes, we introduce a temporal-gradient based initialization method.
This method allocates more Gaussians to regions with high pixel variation over time, thereby focusing the model’s resources on areas with greater temporal complexity, enhancing its overall representational power for complex video content.

Our contributions are summarized as follows:
\begin{itemize}
    \item Our method, GaussianVideo, applies the Gaussian splatting to video representation and video compression for the first time, to the best of our knowledge.
    \item By incorporating a deformable multi-plane-based approach, we achieve efficient video representation and compression.
    \item Our model achieves significantly faster training (5.5x) and rendering (12.5x) speed with lower memory consumption (-78.4\%), maintaining the rendering quality comparable to the state-of-the-art methods.
\end{itemize}
\section{Related Work}
\label{sec:related}
\textbf{Video Compression.}
Traditional video compression methods, such as H.264 \cite{sullivan2012overview} and HEVC \cite{wiegand2003overview}, rely on predictive coding architectures to encode both motion information and residual data from videos.
Neural network-based algorithms \cite{chen2017deepcoder, li2021deep} leverage data-driven approaches for feature extraction and compression.
These early methods, however, have been still tied to conventional hand-crafted approaches,
with intricate connections introduced between numerous sub-components.
Recent methods address the computational inefficiencies by refining traditional codecs \cite{khani2021efficient} and optimizing various parts of the compression pipeline \cite{rippel2021elf}.

\vspace{0.1cm} \noindent
\textbf{Implicit Neural Representation (INR).}
Implicit neural representation is an innovative approach to parametrize diverse signals such as shapes \cite{genova2019learning, genova2020local, gropp2020implicit}, objects \cite{michalkiewicz2019implicit}, or scenes \cite{sitzmann2019scene, jiang2020local}, as neural networks.
Comparing to explicit 3D representations, INR enables continuous and high-resolution representations by approximating a function that directly maps coordinates to signal values \cite{chen2021learning, mehta2021modulated,sitzmann2020implicit}.
Other methods, such as the signed distance function \cite{park2019deepsdf, atzmon2020sal, michalkiewicz2019implicit} or occupancy network \cite{mescheder2019occupancy, peng2020convolutional}, have also been proposed to learn the signal.

\vspace{0.1cm} \noindent
\textbf{Implicit Neural Representation for Videos (NeRV).}
Recently, NeRV \cite{chen2021nerv} has successfully applied INRs to videos.
NeRV is an image-wise implicit representation model that takes a frame index as an input
rather than pixel-wise coordinates.
By generating an entire image per frame instead of individual pixels, NeRV reduces the required sampling rate, which significantly improves encoding and decoding speed.
This approach has shown promising results in video compression tasks as well.

Numerous works have improved NeRV by more effectively leveraging the unique characteristics of video data.
DNeRV \cite{zhao2023dnerv} aims to model videos with large motion by integrating dynamics of adjacent frames and spatial features learned from separate streams.
HNeRV \cite{chen2023hnerv} proposes content-adaptive embedding and a redesigned architecture. SNeRV \cite{kimsnerv} handles various frequency components separately, capturing fine spatial details and motion patterns.

One special aspect about representing a video as a neural network is the fact that then the video compression problem can be reframed as a model compression task, allowing existing model compression and acceleration techniques \cite{frankle2019lottery, gholami2018squeezenext} to be applied to the INR-based networks. 
However, despite the advantages, NeRV-based models fail to provide fast training and real-time rendering yet, since a forward pass through the entire network is needed per sample.

\vspace{0.1cm} \noindent
\textbf{3D Gaussian Splatting.}
Gaussian Splatting \cite{kerbl20233d} offers an innovative explicit method in 3D scene representation.
Unlike traditional volumetric ray-marching \cite{martin2021nerf} or interpolation-based approaches \cite{fridovich2022plenoxels, muller2022instant,xu2022pointnerf}, it directly optimizes 3D Gaussians for efficient, real-time rendering of high-quality views.

While extensive research has been conducted on rendering dynamic scenes in videos using 3D Gaussians with an additional temporal axis (4D) \cite{cao2023hexplane,fridovich2023kplanes,wu20244dgsplat, duan20244d, yang2023real} or compressing images using 2D Gaussians \cite{zhang2025gaussianimage}, studies that model videos by adding a temporal axis to 2D Gaussians remain unexplored.
In this paper, we aim to enhance existing NeRV-based methods in terms of training efficiency and decoding speed by leveraging 2D Gaussian Splatting.

\section{Preliminaries}
\label{sec:preliminaries}

We provide a brief review of the rendering process of Gaussian Splatting \cite{kerbl20233d,zhang2025gaussianimage} in Sec \ref{sec:preliminaries:gauss}, followed by an introduction of the multi-plane based representation in Sec \ref{sec:preliminaries:multiresvox}.

\subsection{Gaussian Splatting}
\label{sec:preliminaries:gauss}

3D Gaussian Splatting \cite{kerbl20233d} is an explicit method for representing 3D scenes.
We define $N$ anisotropic 3D Gaussian ellipsoids, denoted by $G_i(\mathbf{x})$, with a learnable center point $\boldsymbol{\mu}_i \in \mathbb{R}^{3}$ and a learnable covariance matrix $\boldsymbol{\Sigma}_i \in \mathbb{R}^{3 \times 3}$:
\begin{equation}
    G_i(\mathbf{x}) = e^{-\frac{1}{2} (\mathbf{x-\boldsymbol{\mu}_i})^\top \boldsymbol{\Sigma}_i^{-1} (\mathbf{x-\boldsymbol{\mu}_i})},
    \label{eq:gaussian_def}  
\end{equation}
where $\mathbf{x} \in \mathbb{R}^3$ represents a 3D query point and $i = 1, ..., N$.

To render the 3D Gaussian on a 2D image plane, we need a projection process.
As Zwicker \etal~\cite{zwicker2001surface} has demonstrated, the transformed covariance matrix $\bm{\Sigma}'$ in the camera coordinates can be computed using a viewing transformation $\bm{W}$ and the Jacobian $\bm{J}$ of the affine approximation of the projective transformation:
\begin{equation} \label{eq.2}
    \bm{\Sigma}' = \bm{J W \Sigma W^\top J^\top}.
\end{equation}

After determining the ellipse's center position and size from the 2D mean and covariance obtained through the projection, the color $\mathbf{c}_i \in \mathbb{R}^3$ is calculated using spherical harmonic (SH) coefficients, capturing view-dependent color properties based on the camera coordinates.
We subsequently compute the final pixel color $\mathbf{C} \in \mathbb{R}^3$, determined by blending $N$ Gaussians:

\begin{equation}
    \mathbf{C} = \sum_{i=1}^N \mathbf{c}_i \alpha_i \prod_{j=1}^{i-1} (1 - \alpha_j),
    \label{eq:color} 
\end{equation}
where 
$\alpha_i \in \mathbb{R}$ is a weight incorporating the opacity $o_i \in (0, 1)$ and spatial influence of the $i$-th Gaussian, defined as
\begin{equation}
    \alpha_i = o_i e^{-\frac{1}{2} (\mathbf{x} - \mu_i)^\top \Sigma_i^{-1} (\mathbf{x} - \mu_i)}.
    \label{eq:alpha} 
\end{equation}
In this blending process, Gaussians are ordered by their depth values, representing the relative distances of their centers from the camera along the viewing direction.
By accurately accumulating the contributions of all $N$ Gaussians, this process enables precise color blending, producing the final image.

2D Gaussian Splatting \cite{zhang2025gaussianimage} offers an efficient method for representing and compressing an image, enabling minimal GPU memory overhead while significantly enhancing rendering speed.
Unlike 3D Gaussian Splatting, each Gaussians in 2D is defined by a covariance matrix $\boldsymbol{\Sigma} \in \mathbb{R}^{2 \times 2}$ and a center point $\boldsymbol{\mu} \in \mathbb{R}^{2}$.
Since all coordinates reside in the same 2D space, no projection process is required, simplifying the architecture.

Furthermore, this method reduces the number of learnable parameters from 59 in the 3D Gaussian model (which includes parameters for mean, covariance, color from spherical harmonic, projection process, and opacity) to just 8 in the 2D Gaussian model (2 for the mean, 3 for covariance, and 3 for color).
The computation of the color of each pixel using the 2D Gaussian follows a procedure similar to that outlined in Eq.~\eqref{eq:color} and \eqref{eq:alpha}.
Additionally, in the absence of depth and the use of spherical harmonic, the learnable parameters $o_i \in \mathbb{R}$ and $\mathbf{c}_i \in \mathbb{R}^3$ can be merged into a single parameter $c_i \in \mathbb{R}^3$, resulting in a more simplified expression:
\begin{equation}
    \mathbf{C} = \sum_{i=1}^N \mathbf{c}_i e^{-\frac{1}{2} (\mathbf{x} - \mu_i)^T \Sigma_i^{-1} (\mathbf{x} - \mu_i)}.
    \label{eq:rendering_2d} 
\end{equation}

Our framework extends this 2D Gaussian Splatting to the video domain, leveraging its strengths to reduce learnable parameters and enhance computational efficiency.


\subsection{Multi-Plane-based Representation}
\label{sec:preliminaries:multiresvox}

On high-dimensional data such as 4D, MLPs struggle to efficiently capture input details, often requiring deeper networks or larger hidden dimensions, which sharply increases the computational cost and limits scalability.
Multi-plane representation \cite{chan2022eg3d,cao2023hexplane,fridovich2023kplanes} has been widely adopted to address this issue. 
In this approach, pairs of dimensions selected from the input dimension $d$ form $\binom{d}{2}$ explicit planes, where each represents a specific 2D projection of the original space, allowing the model to efficiently capture and represent spatio-temporal relationships within the high-dimensional input.
Each plane belongs to $\mathbb{R}^{m \times n \times c}$, where $m \times n$ is the resolution of the plane and $c$ is the feature dimensionality obtained from each plane.

In the case of 4D inputs, \emph{e.g.}, $\textbf{q} = (x, y, z, t)$, where $x$, $y$, and $z$ represent the 3D spatial coordinates and $t$ represents the temporal axis, points are projected onto 6 different planes: \(XY\), \(XZ\), \(YZ\) for capturing spatial information, and \(XT\), \(YT\), \(ZT\) for spatio-temporal relationships. 
After the projection, the feature $F_{ij}$ for each plane $ij$ is retrieved through bilinear interpolation, and these features are combined using the Hadamard product (element-wise multiplication) $\odot$ to generate the final feature vector:
\begin{equation}
  F(\textbf{q}) = \bigodot_{i,j \in P} F_{ij}(q_i, q_j)
\end{equation}
where $P = \{(x, y), (x, z), (y, z), (x,t), (y,t), (z,t)\}$ denotes
the set of all planes in the 4D inputs, and $q_i$ and $q_j$ represent the coordinates in each dimension of a given plane.

With both the spatial and temporal information of the point $(x, y, z, t)$,
$F(\textbf{q})$ then goes through a lightweight MLP that decodes it into color, density, or other attributes, preserving accurate details at low computational cost.

Multi-plane-based approaches are highly efficient, as their complexity grows only quadratically with increasing input dimensionality, in contrast to the exponential growth seen with naive direct modeling of the space.
It also enables
straightforward access to feature values at specific coordinates by directly storing spatial information on predefined planes, maintaining its efficiency even with a large model.
Furthermore, the factorization across planes allows application of different priors, enhancing the model's representational power.
Multi-scale interpolation can be applied to spatial planes to capture finer spatial details, while temporal priors can be selectively added to space-time planes to better handle dynamic data.

Taking advantage of these benefits, we apply this multi-plane representation to 3D inputs \((x,y,t)\), to efficiently capture spatial and spatio-temporal structures.
This approach enables the model to learn detailed and well-organized information while maintaining computational efficiency.

\begin{figure*}
    \centering
    \includegraphics[width=1\linewidth]{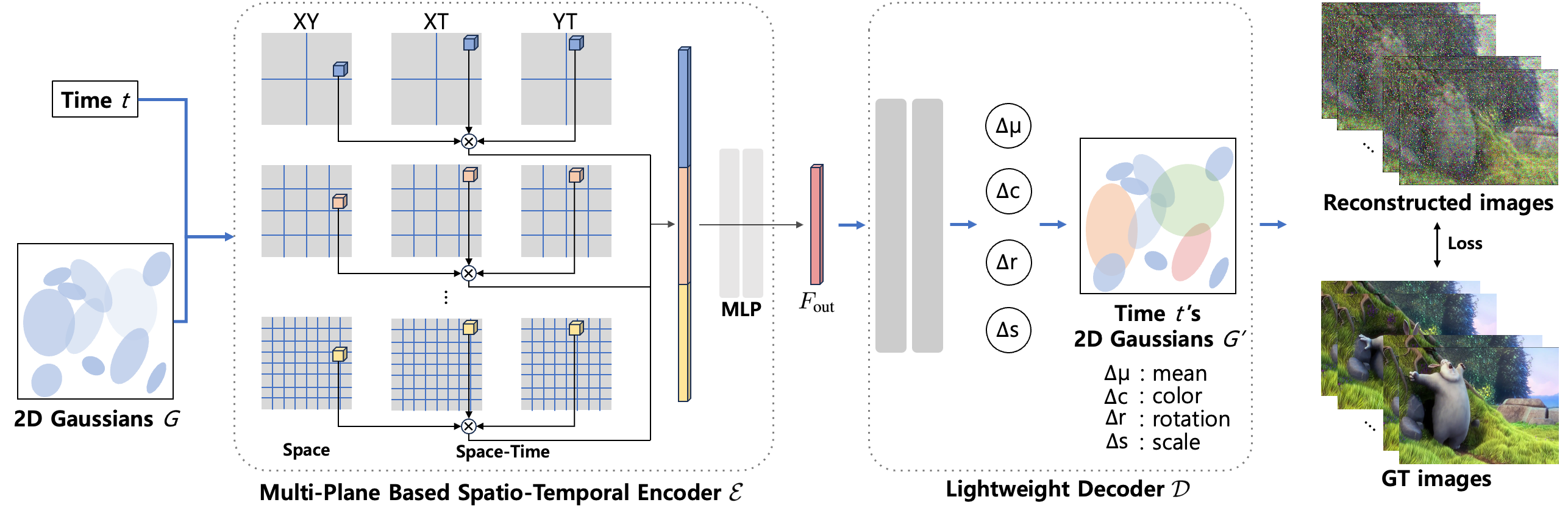}
    \caption{\textbf{Overview of our GaussianVideo.} 
    The encoder $\mathcal{E}$ takes $N$ 2D Gaussians and a time step $t$ as input and produces a feature $F$ fusing the spatio-temporal information.
    From this feature, a lightweight decoder $\mathcal{D}$ reconstructs the deformations for each Gaussian and renders the image.}
    \label{fig:overall_pipeline}
    \vspace{-0.2cm}
\end{figure*}

\section{Method}
\label{sec:method}
As introduced in \cref{fig:overall_pipeline}, our model architecture consists of a spatio-temporal encoder based on multi-planes and a lightweight decoder.
Our framework takes as input the mean values $(x,y)$ of multiple 2D Gaussians along the temporal axis $t$.
The Encoder $\mathcal{E}$ aims to extract a feature across spatio-temporal axes, capturing the essence of the contents in the input video.
The Decoder $\mathcal{D}$ then applies a lightweight MLP on this feature, to predict the color value, 2D mean, and appearance (covariance) that the input Gaussian $(x,y)$ should have at time $t$.
The entire image is reconstructed using these \textit{time-dependent} Gaussians.
At the end, we compute the discrepancy between the ground truth and the reconstructed frame image as the loss, and use it to update the model parameters.
We first describe our encoder and decoder structures (\cref{sec:method:deform}) and our novel temporal gradient-based initialization (\cref{sec:method:init}) in detail, followed by model training (\cref{sec:method:train}).

\subsection{Deformable 2D Gaussian Framework}
\label{sec:method:deform}
The most straightforward approach to represent a video using 2D Gaussians would be having the same number of 2D Gaussians across all frames $T$ in the video, just as done for a single image.
However, this method lacks scalability and becomes inefficient even for a moderately long video.

In a video with sufficiently high frame rates, there is often hugely redundant information between adjacent frames, causing each Gaussian to naturally capture similar information.
Leveraging this, we propose a method that calculates only the \textit{changes}, or \textit{deformations} in each Gaussian, rather than learning new Gaussians for each frame from scratch.

A naive approach to handle this deformation process is to use a deep feed-forward network (MLP).
However, as the number of Gaussians and the length and resolution of the video grow, wider and deeper hidden layers will be needed.
This results in a sharp rise in computational cost and slower processing.
To address this, we choose a hybrid approach that combines a multi-plane-based encoder-decoder structure.
This setup enables us to efficiently extract features corresponding to the input coordinates and decode Gaussian deformations with lower computational overhead.

Specifically, the encoder and decoder are trained to compactly represent (compress) the deformations of the 2D Gaussian’s components, appearance and color.
Formally, given a particular Gaussian $G$ and a time step $t$ as inputs, the encoder $\mathcal{E}$ and decoder $\mathcal{D}$ yield a deformable Gaussian $G'$, which depends on time $t$:
\begin{equation}
    G' = \mathcal{D}(\mathcal{E}(G, t)).
    \label{eq:encoder-decoder}
\end{equation}

\vspace{0.1cm} \noindent
\textbf{Multi-plane based Spatio-Temporal Encoder.}
Given a video $\mathbf{X} \in \mathbb{R}^{T \times H \times W \times 3}$, where $T$ is the number of frames and $H \times W$ is the RGB frame resolution, the encoder $\mathcal{E}$ aims to represent spatio-temporal features for a particular index $t, i, j$, where $1 \le t \le T$, $1 \le i \le H$, and $1 \le j \le W$, denoted by $F(t, i, j)\in \mathbb{R}^C$, where $C$ is the feature dimensionality.

Instead of treating the video as a 3D tensor as is, we
adopt the 3D multi-plane \cite{fridovich2023kplanes,cao2023hexplane} in our deformable encoder, for the sake of efficiency.
From the three axes, we have $\binom{3}{2} = 3$ explicit planes:
a spatial $XY$ plane ($\in \mathbb{R}^{H' \times W' \times C}$) and two spatio-temporal planes, $XT$ ($\in \mathbb{R}^{H' \times T' \times C}$), $YT$ ($\in \mathbb{R}^{W' \times T' \times C}$), where 
$C$ is the feature dimensionality for the coordinates on each plane.
Here, $H' \ll H$, $W' \ll W$, and $T' \ll T$ represent the sizes of each axis on the corresponding planes, respectively.

Each 2D plane learns spatio-temporal patterns corresponding to their axes independently, and we take Hadamard product ($\circ$) of these features across planes to aggregate them back to 3D:
\begin{equation}
  F(\textbf{q}) = F_{XY}(q_x, q_y) \circ F_{XT}(q_x, q_t) \circ F_{YT}(q_y, q_t),
\end{equation}
where $F_{XY}$, $F_{XT}$, and $F_{YT}$ represent the features from each respective plane.
$\textbf{q}=(x,y,t)$ denotes the input coordinate, which corresponds to the 2D Gaussian's mean value $(x,y)$ and a specific time $t$.
$q_x$, $q_y$, and $q_t$ are the projections of $\textbf{q}$ onto the respective axes for each plane.

By performing this process at $R$ different resolutions $\{r_i: i = 1, ..., R\}$, we obtain a set of multi-resolution features.
Concatenating these features and passing through an MLP layer yields the final feature $F_\text{out} \in \mathbb{R}^h$ that incorporates multi-resolution spatio-temporal information:
\begin{equation}
  F_\text{out} = \text{MLP}(\text{concat}(F^{r_1}, F^{r_2}, \dots, F^{r_R}))
\end{equation}
where each $F^{r_i}$ is the feature obtained at resolution $r_i$.
This final representation, $F_\text{out}$, is the output of our encoder $\mathcal{E}$, and is trained to reconstruct the original image at time $t$ through the decoder, explained below.

\vspace{0.1cm} \noindent
\textbf{Lightweight Decoder.}
From the encoding $F_\text{out} \in \mathbb{R}^h$, the decoder $\mathcal{D}$ aims to estimate the variations in each Gaussian's attributes, such as color, rotation, and scale.
Formally, the process can be written as
\begin{equation}
  G' = \mathcal{D}(F_\text{out})
\end{equation}
where $G'$ denotes the variations in Gaussian attributes (\textit{e.g.}, color, rotation, scale).
The decoder $\mathcal{D}$ can be as simple as a single-layer MLP, because $F_\text{out}$ already encapsulates a substantial amount of spatial and spatio-temporal information.
We reconstruct the entire image at $t$ through the rendering process in \cref{eq:rendering_2d}, using these $N$ deformed Gaussians $G'$.


\subsection{Temporal Gradient-based Initialization}
\label{sec:method:init}

According to RAIN-GS \cite{jung2024relaxing}, properly initializing the mean values of the Gaussians is crucial for convergence.
In case of videos, some spatial or temporal regions tend to have more dynamic motions than others (\emph{e.g.}, background).
As the dynamic regions have higher RGB variance over time, it is intuitive to assign more Gaussians for a more accurate representation in these regions.
Thus, we initialize Gaussians proportional to the distribution of spatio-temporal changes by leveraging \textit{temporal gradients} from the video.
The temporal gradient represents the pixel-wise difference between two consecutive frames, capturing changes directly without the need for complex external models.

To calculate the temporal gradient, we take two consecutive frames, $I_t$ and $I_{t+1}$, and compute the absolute pixel-wise difference $\Delta I_t = \left| I_{t+1} - I_t \right|$ across all consecutive frames with $t = 1, ..., T-1$.
Then, we sum these gradients over the entire video, yielding a cumulative map $\mathbf{M} \in \mathbb{R}^{H \times W \times 3}$ of temporal variations:
\begin{equation}
  \mathbf{M} = \sum_{t=1}^{T-1} \Delta I_t
\end{equation}
where its each element reflects the extent of change throughout the video.

\begin{figure}
    \centering
    \includegraphics[width=1\linewidth]{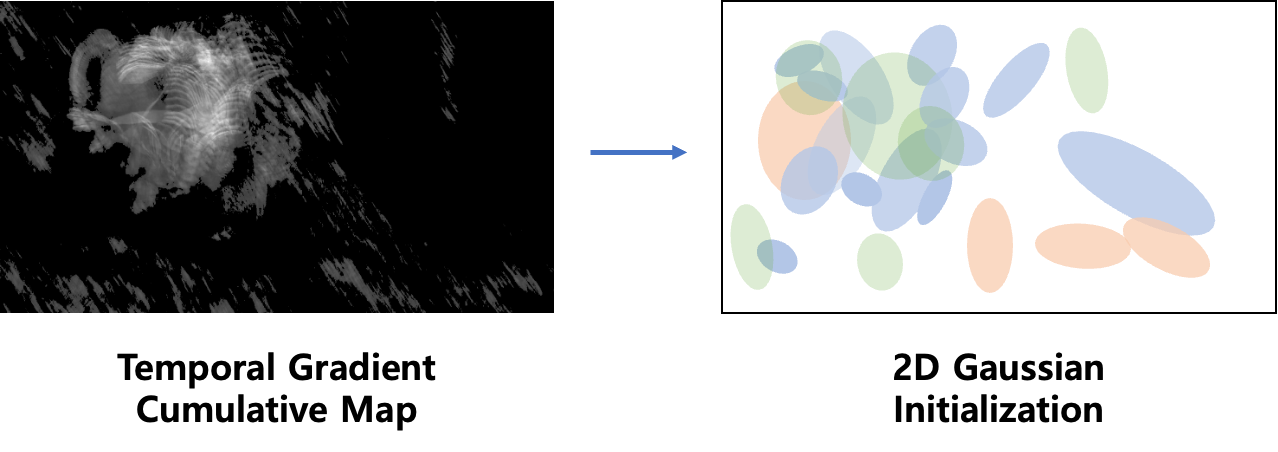}
    \vspace{-0.6cm}
    \caption{\textbf{Temporal gradient-based Initialization}. Brighter areas indicate regions with higher variation. We initialize more Gaussians in these brighter regions.}
    \label{fig:temporal}
\end{figure}

As illustrated in \cref{fig:temporal}, using this cumulative temporal variation map, we prioritize regions with higher variation by initializing more Gaussians in them.
This process has a negligible computational cost and is performed only once before training.
Such an initialization enables our model to better capture dynamic regions in the video, enhancing reconstruction quality in the areas with significant changes.


\subsection{Model Training}
\label{sec:method:train}

Our method takes the mean $(x, y)$ of 2D Gaussians and a time step $t$ as inputs, and infers the color and appearance of the Gaussian at the specified time $t$.
Using these 2D Gaussians dependent on $t$, we render an image $\hat{I}$ that approximates the target frame at $t$, over the entire video with $1 \le t \le T$.
To train our model, we minimize the $L_2$ reconstruction loss $\mathcal{L}_{\text{recon}}$ between the rendered image $\hat{I}$ and the true image $I$:
\begin{equation}
  \mathcal{L}_{\text{recon}} = \frac{1}{HW} \sum_{x=1}^{H} \sum_{y=1}^{W} \left( \hat{I}(x, y) - I(x, y) \right)^2
\end{equation}
where $H$ and $W$ are the height and width of the image, and $\hat{I}(x, y)$ and $I(x, y)$ represent the color values of the rendered and target images at pixel $(x, y)$, respectively.

In addition, we also apply the Total Variation (TV) Loss $\mathcal{L}_{\text{TV}}$, which is commonly used in multi-plane-based methods \cite{wu20244dgsplat,cao2023hexplane,fridovich2023kplanes,fang2022fast,sun2022direct}, to encourage smoothness in each plane grid within the encoder.
The TV Loss is defined as
\begin{equation}
  \mathcal{L}_{\text{TV}} = \sum_{i, j} \left( \left| F(i+1, j) - F(i, j) \right| + \left| F(i, j+1) - F(i, j) \right| \right) \nonumber
\end{equation}
where $F(i, j)$ represents the feature values in the encoder’s plane grid at $(i, j)$.
This loss helps to regularize changes over time, keeping it smooth and preventing sharp changes.

The overall loss $\mathcal{L}$ linearly combines the two losses:
\begin{equation}
  \mathcal{L} = \mathcal{L}_{\text{recon}} + \lambda \mathcal{L}_{\text{TV}}
\end{equation}
where $\lambda$ is a hyper-parameter to control relative strength of each loss. With this loss setting, our model is trained to accurately render color while maintaining smoothness in the encoded spatial structure.


\section{Experiments}
\label{sec:exp}
\subsection{Experimental Settings}

\begin{figure*}
    \centering
    \includegraphics[width=1\linewidth]{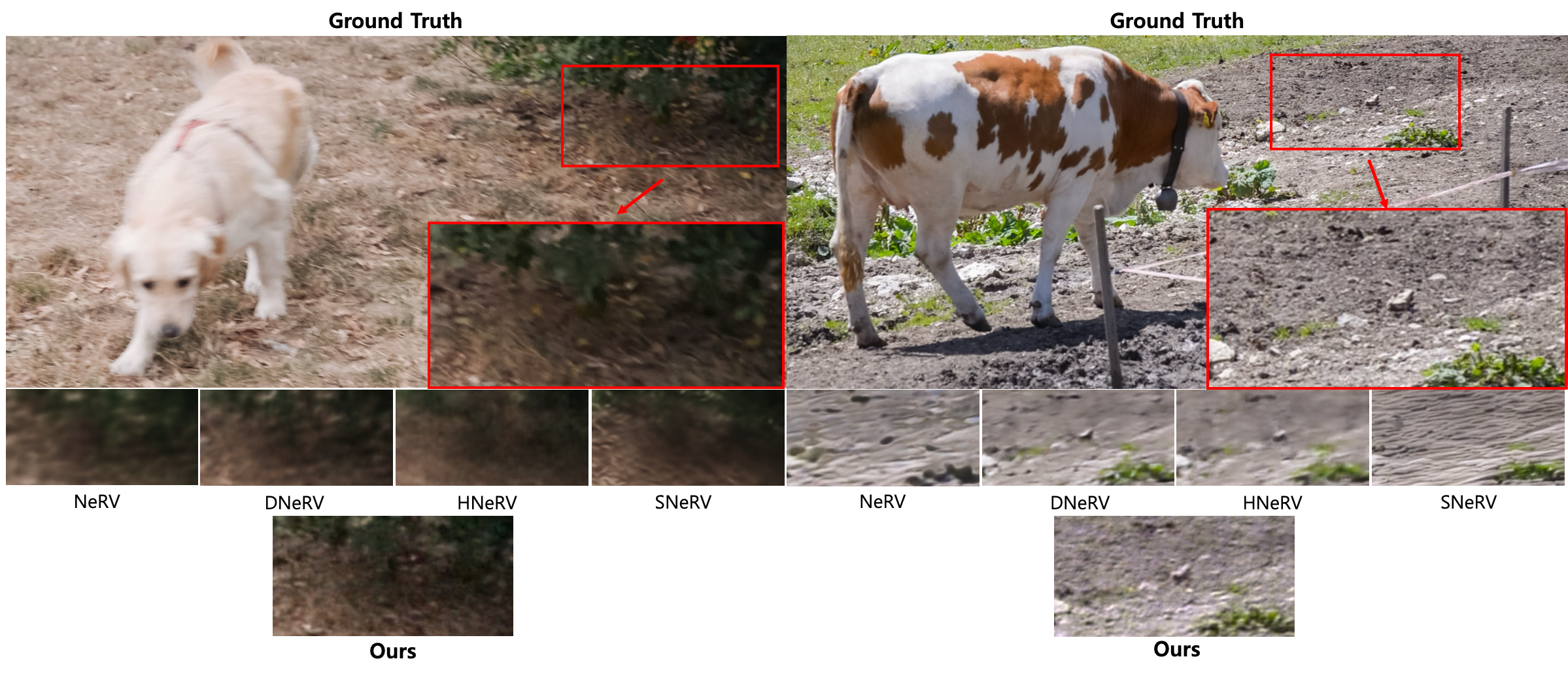}
    \vspace{-0.6cm}
    \caption{\textbf{A Qualitative Example on DAVIS}. This example is from all models with size 0.75M.}
    \label{fig:qualitative}
\end{figure*}


\vspace{0.1cm} \noindent
\textbf{Datasets.}
We conduct experiments on three high-resolution datasets: Bunny \cite{roosendaal2008big}, DAVIS \cite{perazzi2016benchmark}, and UVG \cite{mercat2020uvg}.
The Bunny Video has a resolution of 720 $\times$ 1280 and consists of 132 frames.
For our experiments, we apply a center crop and resize it to 640 $\times$ 1280. 
From the DAVIS dataset, we select 10 videos with a resolution of 960 $\times$ 1920.
The UVG dataset consists of 7 videos, also at 960 $\times$ 1920 resolution, with total lengths of either 300 or 600 frames.

\vspace{0.1cm} \noindent
\textbf{Baselines.}
We compare with the state-of-the-art NeRV model \cite{chen2021nerv} and its variations SNeRV \cite{kimsnerv}, HNeRV \cite{chen2023hnerv}, and
DNeRV \cite{zhao2023dnerv} for video representation and compression.

\vspace{0.1cm} \noindent
\textbf{Evaluation Metrics.}
For video reconstruction quality, we report two widely-used metrics, peak-signal-to-noise ratio (PSNR) and multiscale structural similarity index (MS-SSIM) \cite{wang2004image}.
For video compression efficiency, we measure training time, decoding throughput in frame-per-second (FPS), VRAM usage, and bits per pixel (BPP), which reflects compression efficiency by measuring the bits needed per pixel while balancing visual quality.

\vspace{0.1cm} \noindent
\textbf{Implementation Details.}
We implement our method using PyTorch \cite{paszke2019pytorch} and conduct all experiments on a single A5000 GPU.
During training on the Bunny Dataset with a model size of 0.35M, we set the resolution for the spatial axes \(x, y\) to 16 and the temporal axis \(t\) to 8. Two multi-resolution planes are used with scaling ratios of 1 and 2. 
We set the total number of Gaussians to 31,024.
For the 0.75M model, the multi-resolution configuration is identical to the 0.35M model. The spatial resolution for is set to 32, and the temporal resolution is doubled to 16, resulting in 50,034 Gaussians. For the 1.5M model, three multi-resolution planes with scaling ratios of 1,2, and 4 are employed. The spatial resolution 32, while temporal resolution is set to 8. This configuration results in 55,734 Gaussians. Finally, for the 3.0M model, three multi-resolution planes with scaling ratios of 1, 2, and 4 are used, similar to the 1.5M model. The spatial resolution is increased to 48, and the temporal resolution to 12, resulting in a total of 81,954 Gaussians. In all processes, we train the parameters using the Adan \cite{xie2024adan} optimizer with a learning rate 0.001. Further details are provided in the Appendix.

\begin{table}
  \centering \scriptsize \renewcommand{\tabcolsep}{3pt}
  \begin{tabular}{c|l|cc|c|rc}
  \toprule
  \textbf{Size} & \textbf{Model} & \textbf{PSNR}$\uparrow$ & \textbf{MS-SSIM}$\uparrow$ & \textbf{Training}$\downarrow$ & \textbf{FPS$\uparrow$} & \textbf{GPU Mem$\downarrow$} \\
  \midrule
   & NeRV \cite{chen2021nerv}   & 26.5  & 0.854  & \underline{13 min}  &  \underline{340}   & 1.99GB   \\
   & DNeRV \cite{zhao2023dnerv}   &  \underline{30.6}  & 0.926  & 55 min  & 75   & 2.72GB   \\
  0.35M & HNeRV \cite{chen2023hnerv}  & 30.0  & 0.915  & 24 min   & 215   & 2.32GB  \\
   & SNeRV \cite{kimsnerv}  &  \textbf{31.1}  &  \textbf{0.959}  & 49 min   & 80  &  \underline{1.64GB}  \\
   & \textbf{Ours}  &  30.4  &  \underline{0.935}   &  \textbf{11 min}   &  \textbf{584}  &  \textbf{1.20GB}  \\
   \midrule
   & NeRV \cite{chen2021nerv}   & 28.2  & 0.904  & \underline{13 min}  & \underline{333}   & 1.99GB   \\
   & DNeRV \cite{zhao2023dnerv}   &  \underline{33.4}  &  \underline{0.962}  & 55 min  & 47   & \textbf{1.43GB}   \\
  0.75M & HNeRV \cite{chen2023hnerv}  & 32.8  & 0.956  & 25 min   & 225   & 3.25GB  \\
   & SNeRV \cite{kimsnerv}  & \textbf{34.6}  & \textbf{0.980}  & 50 min   & 78  & 2.43GB  \\
   & \textbf{Ours}  & 32.2  &   0.953  &  \textbf{12 min}   & \textbf{545}  & \underline{1.70GB}  \\
   \midrule
   & NeRV \cite{chen2021nerv}   & 30.8  & 0.950  & \underline{14 min}  & \underline{337}   & 2.44GB   \\
   & DNeRV \cite{zhao2023dnerv}   & 33.3  & 0.961  & 58 min  & 38   & \textbf{1.56GB}   \\
  1.5M & HNeRV \cite{chen2023hnerv}  & \underline{35.6}  & \underline{0.977}  & 28 min   & 174   & 4.52GB  \\
   & SNeRV \cite{kimsnerv}  & \textbf{37.7}  & \textbf{0.989}  & 70 min   & 45  & 4.34GB  \\
   & \textbf{Ours}  & 33.0  &   0.960  & \textbf{12 min}   & \textbf{512}  & \underline{1.79GB}  \\
   \midrule
  & NeRV \cite{chen2021nerv}   & 34.2  & 0.977  & \underline{14 min}  & \underline{323}   & 4.08GB   \\
   & DNeRV \cite{zhao2023dnerv}   & 37.0  & 0.984  & 68 min  & 27   & \textbf{1.78GB}   \\
  3.0M & HNeRV \cite{chen2023hnerv}  & \underline{38.0}  & \underline{0.988}  & 27 min   & 96   & 8.33GB  \\
   & SNeRV \cite{kimsnerv}  & \textbf{40.6}  & \textbf{0.994}  & 72 min   & 38  & 8.77GB  \\
   & \textbf{Ours}  & 34.8  &   0.979  & \textbf{13 min}   & \textbf{475}  & \underline{1.89GB}  \\
  \bottomrule
  \end{tabular}
  \vspace{-0.2cm}
  \caption{\textbf{Overall performance comparison on the Bunny Dataset}. PSNR and MS-SSIM measure the image quality, and FPS and GPU Memory are related to rendering efficiency. For a fair comparison, we set the same model size for other NeRV-like models. Specifically, model size refers to the number of parameters required to reconstruct a video, including the parameters of the Encoder, Decoder, and Gaussians in our case.}
  \label{exp:bunny}
\end{table}

\subsection{Results and Discussion}

\noindent
\textbf{Video Reconstruction.}
\cref{exp:bunny} compares the performance of various models evaluated on the Bunny Dataset, which has a resolution of 640 $\times$ 1280.
We assess models of different sizes after training them for 300 epochs.
The results indicate that GaussianVideo achieves 71.7\% faster rendering speed of 584 FPS than the second best model (NeRV), while keeping a competitive PSNR of 30.4 with a model size of 0.35M.
Another notable aspect of our method is that the rendering speed is not significantly affected even with a 10 times larger model size, dropping from 584 to 475 FPS (18.6\%).
Other models, on the other hand, exhibit a significant drop (64.0\% for DNeRV, 46.5\% for DNeRV, 52.5\% for SNeRV) in throughput and a substantial increase in GPU memory consumption as the model size grows from 0.35M to 3.0M.
As the cost of this competent throughput, our method exhibits less improvement in the image quality though.
However, it still achieves superior quality and efficiency compared to the vanilla NeRV.
From this result, we conclude that our GaussianVideo improves the NeRV both in quality and efficiency, but putting more emphasis on efficiency compared to other recent innovations (DNeRV, HNeRV, and SNeRV).
On the DAVIS and UVG datasets with 960 $\times$ 1920 resolution, we aim to evaluate scalability.
As shown in \cref{exp:davis} and \ref{exp:uvg}, our model renders well even on this high resolution input video.
The qualitative results, presented in \cref{fig:qualitative}, demonstrate that our model captures fine-grained details, such as pebbles or grass, better than other models.
Unlike most NeRV-based methods, which lose intricate details during the compression and reconstruction process due to embedding vector reduction, our model preserves these details by leveraging 2D Gaussians reponsible for regions near each pixel, ensuring more accurate reconstruction.

\begin{table*}
\centering \footnotesize
\begin{tabular}{l|cccccccccccc}
\toprule
\textbf{960 $\times$ 1920}  & \textbf{B-swan} & \textbf{B-trees} & \textbf{Boat} & \textbf{B-dance} & \textbf{Camel} & \textbf{C-round} & \textbf{C-shadow} & \textbf{Cows} & \textbf{Dance} & \textbf{Dog} & \textbf{Avg.} \\
\midrule
NeRV \cite{chen2021nerv} & 24.41 & 23.80 & 28.08 & 22.03 & 21.82 & 20.93 & 23.88 & 20.43 & 24.56 & 26.69 & 23.66  \\
DNeRV \cite{zhao2023dnerv} & 27.22 & \underline{26.34} & 30.70 & \textbf{26.51} & \underline{24.43} & \underline{25.25} & \underline{27.87} & \underline{22.66} & 27.48 & 28.04 & 26.65  \\
HNeRV \cite{chen2023hnerv} & 25.60 & 24.71 & 30.27 & 25.79 & 23.25 & 23.24 & 26.31 & 21.95 & 26.45 & 26.91 & 25.49  \\
SNeRV \cite{kimsnerv} & \textbf{28.14} & \textbf{26.73} & \textbf{32.19} & 25.42 & \textbf{24.63} & \textbf{25.37} & \textbf{30.17} & 22.25 & \textbf{28.40} & \underline{29.51} & \textbf{27.28}  \\
\midrule
\textbf{Ours} & \underline{27.99} & 26.23 & \underline{31.40} & \underline{26.12} & 23.69 & 23.58 & 26.89 & \textbf{23.57} & \underline{27.56} & \textbf{29.55} & \underline{26.66} \\
\bottomrule
\end{tabular}%
\vspace{-0.2cm}\caption{\textbf{PSNR on DAVIS in 960 $\times$ 1920}. The first and second best models are \textbf{boldfaced} and \underline{underlined}, respectively.}
\label{exp:davis}
\end{table*}

\begin{figure}
    \centering
    \includegraphics[width=1\linewidth]{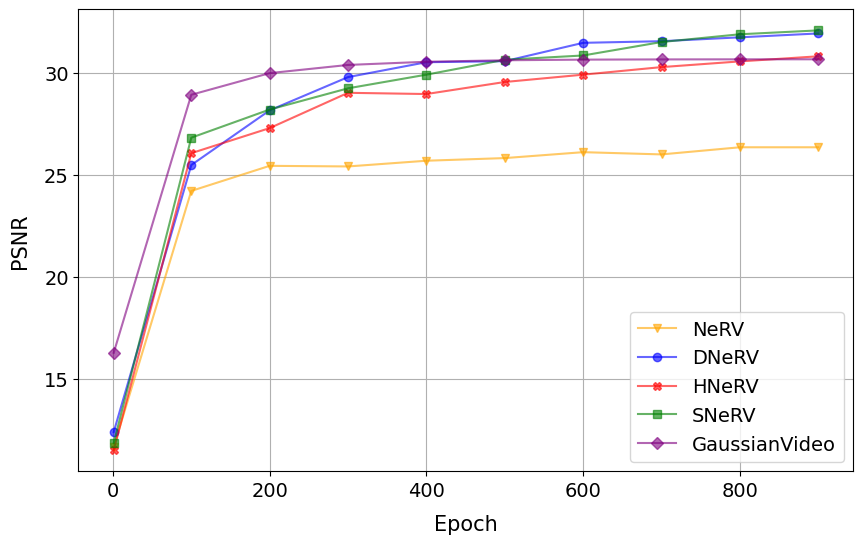}
    \vspace{-0.6cm}
    \caption{\textbf{PSNR results over Epochs.} A models with a size of 0.35M are trained over 900 epochs.}
    \label{fig:psnr900epochs}
\end{figure}

\cref{fig:psnr900epochs} illustrates the PSNR values across the training epochs for the 0.35M model, showing that GaussianVideo converges faster than other models.
Furthermore, since our model requires significantly less time per epoch compared to other models, it reaches the final PSNR much faster.

\begin{figure}
    \centering
    \begin{subfigure}[b]{0.48\linewidth}
        \centering
        \includegraphics[width=\linewidth]{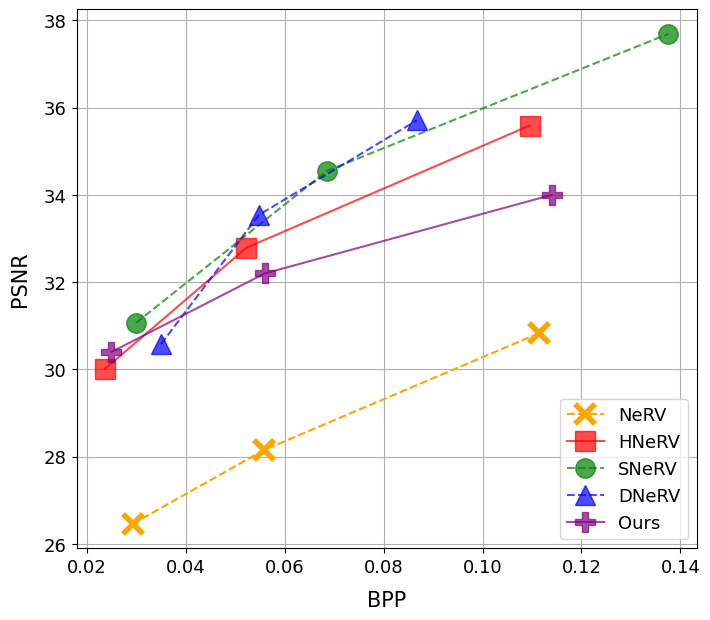}
        \caption{BPP \textit{vs}. PSNR}
        \label{fig:bpp_psnr_bunny}
    \end{subfigure}
    \hfill
    \begin{subfigure}[b]{0.49\linewidth}
        \centering
        \includegraphics[width=\linewidth]{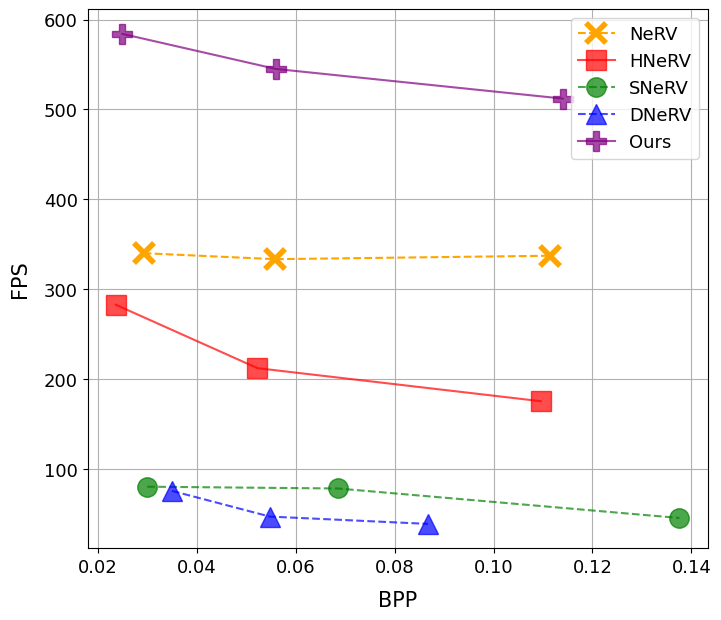}
        \caption{BPP \textit{vs}. FPS}
        \label{fig:bpp_fps_bunny}
    \end{subfigure}
    \vspace{-0.2cm}
    \caption{\textbf{Compression results} on Bunny Dataset.}
    \label{fig:combined_bpp_plots}
\end{figure}

\noindent
\textbf{Video Compression.} Unlike other models, we conduct experiments using only an 8-bit model quantization without applying pruning or entropy encoding.
We compare video compression performance of all competing models in \cref{fig:combined_bpp_plots}.
The results indicate that our model achieves competitive quality in PSNR (\cref{fig:bpp_psnr_bunny}), while maintaining significantly higher rendering speed, even as BPP increases (\cref{fig:bpp_fps_bunny}), outperforming other models in both aspects.

\begin{table}
\centering \footnotesize
\begin{tabular}{l|ccccccccc}
\toprule
\textbf{960 $\times$ 1920} & \textbf{Avg. PSNR$\uparrow$} & \textbf{Avg. Training$\downarrow$} & \textbf{Avg. FPS$\uparrow$} \\
\midrule
NeRV \cite{chen2021nerv} & 28.00 & \underline{139 min} & \underline{228} \\
DNeRV \cite{zhao2023dnerv} & \textbf{31.99} & 683 min & 51 \\
HNeRV \cite{chen2023hnerv} & 30.15 & 140 min & 108 \\
SNeRV \cite{kimsnerv} & \underline{30.80} & 482 min & 34 \\
\midrule
\textbf{Ours} & 29.51 & \textbf{75 min} & \textbf{287} \\
\bottomrule
\end{tabular}%
\vspace{-0.2cm}
\caption{\textbf{PSNR on UVG in 960 $\times$ 1920}. The first and second best models are \textbf{boldfaced} and \underline{underlined}, respectively.}
\label{exp:uvg}
\end{table}

\subsection{Ablation Studies}

\vspace{0.1cm} \noindent
\textbf{Deformable Gaussians.}
First, we verify the effect of our multi-plane-based deformable Gaussian (\cref{sec:method:deform}) against a feed-forward network (MLP).
For a fair comparison, we match the number of learnable parameters for both.
As shown in \cref{ablation:deform},
our multi-plane-based approach performs significantly better than the MLP-based method, both in image quality (PSNR) and in efficiency (training time and throughput in FPS).

\begin{table}[h!]
\centering \footnotesize
\begin{tabular}{l|ccr}
    \toprule
    \textbf{Deform. method} & \textbf{PSNR (dB)$\uparrow$} & \textbf{Training$\downarrow$} & \textbf{FPS$\uparrow$} \\
    \midrule
    MLP & 28.4 & 62 min & 30 \\
    Multi-plane-based (Ours) & \textbf{34.8} & \textbf{13 min} & \textbf{475} \\
    \bottomrule
\end{tabular}%
\vspace{-0.2cm}
\caption{\textbf{Ablation study on deformation methods}. Compared to the MLP-based, our multi-plane approach produces  significantly better images (PSNR) more  efficiently (Training time and FPS).}
\label{ablation:deform}
\end{table}
\begin{table}
\centering \footnotesize
\begin{tabular}{l|cc}
    \toprule
    \textbf{Init. method} & \textbf{PSNR (dB)$\uparrow$} & \textbf{FPS$\uparrow$} \\
    \midrule
    Random initialization & 29.5 & - \\
    Optical flow & 30.2 & 0.79 \\
    Temporal gradient (Ours) & \textbf{30.4} & \textbf{106} \\
    \bottomrule
\end{tabular}%
\vspace{-0.2cm}
\caption{\textbf{Ablation study on Gaussian initialization}. We compare the random initialization, the optical flow-based method, and the temporal gradient method, evaluating their PSNR and FPS, where FPS refers to the processing speed in terms of frames per second.}
\label{ablation:init}
\end{table}

\vspace{0.1cm} \noindent
\textbf{Initialization.}
We compare three initialization methods: 1) randomly initializing within the image resolution range, 2) using an external model to extract optical flow to identify and sample pixels with a significant motion, and 3) our approach using the temporal gradient (\cref{sec:method:init}).
From the result in \cref{ablation:init}, both optical flow and ours, taking the motion dynamics into account to allocate Gaussians, achieve superior PSNR over random initialization.
Our approach, however, renders the video 134 times faster than optical flow, more suitable for video compression, where real-time decoding is important.

\section{Summary and Limitations}
\label{sec:conclusion}

In this work, we present a novel framework for video representation and compression using deformable 2D Gaussian splatting, effectively capturing temporal changes by conditioning Gaussian properties on the time steps.
Our multi-plane based encoder-decoder structure enhances scalability, enabling high-dimensional video modeling with reduced computational overhead. 
Furthermore, we introduce a temporal-gradient-based initialization, allocating resources to regions with high temporal variation, which strengthens the model’s adaptability and overall representational capacity.
According to our experiments, our method demonstrates competitive reconstruction quality, while offering significant advantages in training time and decoding speed.

\vspace{0.1cm} \noindent
\textbf{Limitations.}
While GaussianVideo demonstrates excellent performance, there are areas for improvement.
First, the resolution of the plane of encoder and the number of multi-scales need to be specified for each video.
It would be an interesting future direction to automate this process by adapting these parameters based on the information density within the video.
Second, although our experiments have shown that the current implementation is already highly efficient, further low-level optimization might lead to even higher throughput.

\clearpage
{
  \small
  \bibliographystyle{ieeenat_fullname}
  \bibliography{main}
}

\clearpage
\setcounter{page}{1}
\maketitlesupplementary

\pagenumbering{roman}
\renewcommand\thesection{\Alph{section}}
\renewcommand\thetable{\Roman{table}}
\renewcommand\thefigure{\Roman{figure}}
\setcounter{section}{0}
\setcounter{table}{0}
\setcounter{figure}{0}

\section{Appendix}
\label{appendix}


\subsection{Experimental Setting}
\label{appendix:exp}

\vspace{0.1cm} \noindent
\textbf{Datasets.}
We evaluate our method on two video datasets: DAVIS and UVG, as summarized in \cref{apdx:video_datasets}. DAVIS consists of various video sequences with a resolution of \(1920 \times 960\) and frame counts ranging from 40 to 104. For instance, sequences such as \emph{Blackswan} and \emph{Cows} contain 50 and 104 frames, respectively. UVG, on the other hand, provides longer video sequences with a consistent resolution of \(1920 \times 960\) and up to 600 frames, as seen in examples like \emph{Beauty} and \emph{ReadySteadyGo}. These datasets offer a diverse range of content for evaluating the scalability and performance of our method.

\begin{table}[ht]
    \centering \footnotesize
    \begin{tabular}{llcc}
        \toprule
        \textbf{Dataset} & \textbf{Video} & \textbf{Resolution} & \textbf{Number of Frames} \\
        \midrule
        \multirow{10}{*}{\textbf{DAVIS}} 
        & Blackswan & 1920 x 960 & 50 \\
        & Bmx-trees & 1920 x 960 & 80 \\
        & Boat & 1920 x 960 & 75 \\
        & Breakdance & 1920 x 960 & 84 \\
        & Camel & 1920 x 960 & 90 \\
        & Car-roundabout & 1920 x 960 & 75 \\
        & Car-shadow & 1920 x 960 & 40 \\
        & Cows & 1920 x 960 & 104 \\
        & Dancejump & 1920 x 960 & 60 \\
        & Dog & 1920 x 960 & 60 \\
        \midrule
        \multirow{7}{*}{\textbf{UVG}}
        & Beauty & 1920 x 960 & 600 \\
        & Bosphorus & 1920 x 960 & 600 \\
        & HoneyBee & 1920 x 960 & 600 \\
        & Jockey & 1920 x 960 & 600 \\
        & ReadySteadyGo & 1920 x 960 & 600 \\
        & ShakeNDry & 1920 x 960 & 300 \\
        & YachtRide & 1920 x 960 & 600 \\
        \bottomrule
    \end{tabular}
    \caption{Video datasets with resolution and number of frames.}
    \label{apdx:video_datasets}
\end{table}

\vspace{0.1cm} \noindent
\textbf{Implementation Details.}
The details of hyperparameters are summarized in \cref{apdx:hyper}. For videos with a resolution of \(640 \times 1280\), the \(0.35M\) model uses 31,024 Gaussians with resolutions of \(16 \times 16\) for the \(x, y\) axes and 8 for \(t\). Two multi-resolution planes are used with scaling ratios of \(1\) and \(2\). Enlarging the model size to \(0.75M\), the number of Gaussians rises to 50,034, and the resolutions are set to \(32 \times 32\) for \(x, y\) and 16 for \(t\), maintaining the same multi-resolution configuration. For the \(1.5M\) model, we use 55,734 Gaussians with resolutions of \(32 \times 32\) for \(x, y\) and 8 for \(t\), extending to three multi-resolution planes with ratios of \(1, 2, 4\). Finally, the \(3.0M\) model uses 81,954 Gaussians with resolutions of \(48 \times 48\) for \(x, y\) and 12 for \(t\), retaining the same multi-resolution configuration as the \(1.5M\) model.

For videos with a resolution of \(960 \times 1920\), the \(0.35M\) model is configured with 38,704 Gaussians and resolutions of \(8 \times 8\) for \(x, y\) and 4 for \(t\), with two multi-resolution planes using scaling ratios of \(1\) and \(2\). The \(0.75M\) model increases the number of Gaussians to 60,544 and resolutions to \(32 \times 32\) for \(x, y\) and 8 for \(t\). For the \(1.5M\) model, we configure it with 68,070 Gaussians and resolutions of \(32 \times 32\) for \(x, y\) and 6 for \(t\), using three multi-resolution planes with ratios of \(1, 2, 4\). Lastly, the \(3.0M\) model employs 99,666 Gaussians with resolutions of \(48 \times 48\) for \(x, y\) and 10 for \(t\), maintaining the same multi-resolution configuration as the \(1.5M\) model.

As the video size increases, the number of Gaussians is scaled accordingly to effectively handle the higher resolution. By increasing the number of Gaussians and their respective resolutions, our method is able to capture the additional spatio-temporal details introduced by larger video dimensions, ensuring robust performance across varying resolutions.

\begin{table}[ht]
    \centering \footnotesize
    \begin{tabular}{@{}c|ccccccc@{}}
        \toprule
        \textbf{Video size} & \textbf{size (M)} & \textbf{Num of $G$}& $\boldsymbol{x}$ & $\boldsymbol{y}$ & $\boldsymbol{t}$ & \textbf{ratio} & \\ 
        \midrule
        $640 \times 1280$ & 0.35 & 31,024 & 16 & 16 & 8 & 1,2 \\
        $640 \times 1280$ & 0.75 & 50,034 & 32 & 32 & 16 & 1,2 \\
        $640 \times 1280$ & 1.5 & 55,734 & 32 & 32 & 8 & 1,2,4 \\
        $640 \times 1280$ & 3 & 81,954 & 48 & 48 & 12 & 1,2,4 \\
        $960 \times 1920$ & 0.35 & 38,704 & 8 & 8 & 4 & 1,2 \\
        $960 \times 1920$ & 0.75 & 60,544 & 32 & 32 & 8 & 1,2 \\
        $960 \times 1920$ & 1.5 & 68,070 & 32 & 32 & 6 & 1,2,4 \\
        $960 \times 1920$ & 3 & 99,666 & 48 & 48 & 10 & 1,2,4 \\
        \bottomrule
    \end{tabular}
    \caption{\textbf{GaussianVideo} architecture details.}
    \label{apdx:hyper}
\end{table}

\subsection{More Qualitative Results}
\label{appendix:qual}

\begin{figure*}
    \centering
    \includegraphics[width=1\linewidth]{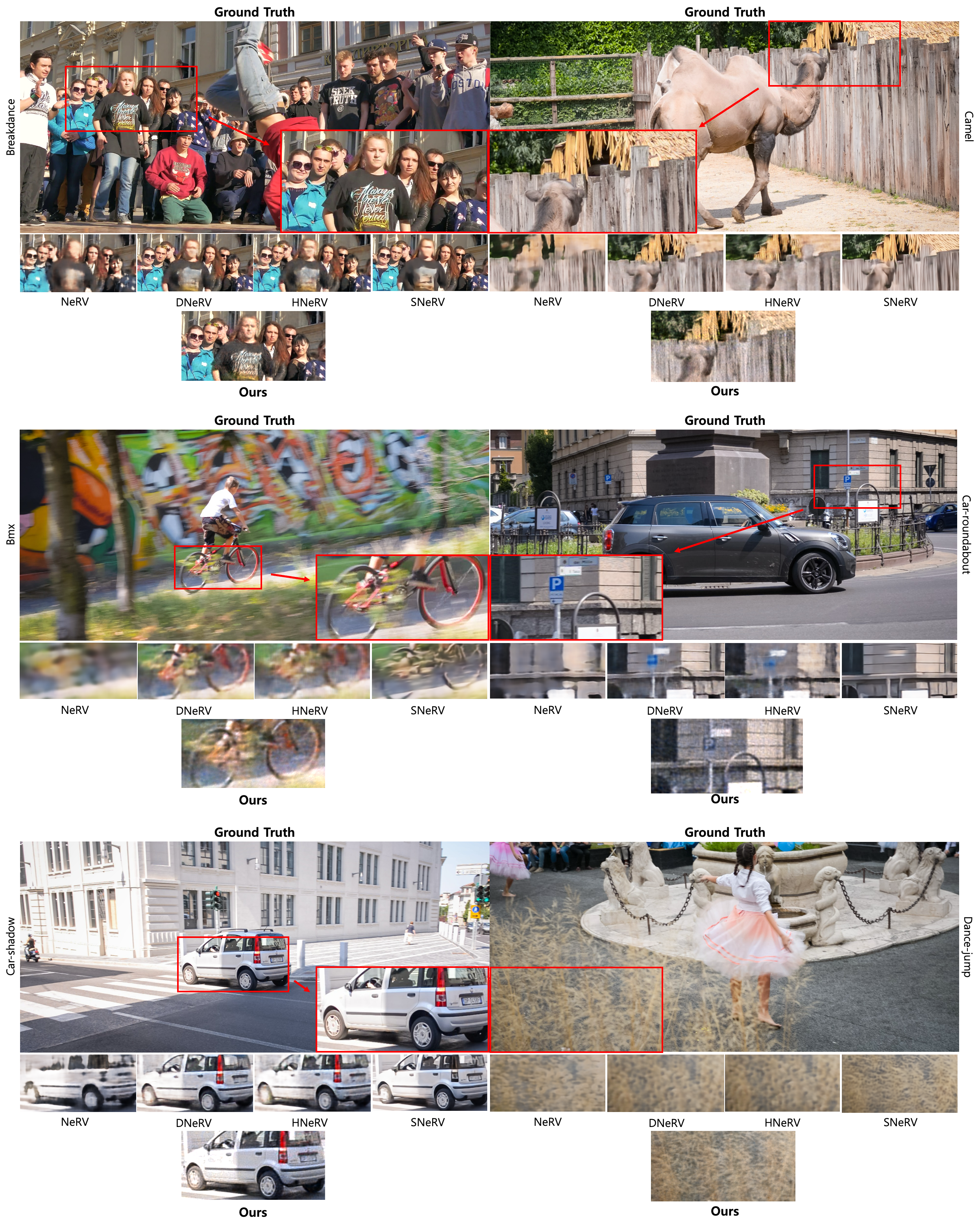}
    \caption{Qualitative comparison of different models on videos from the DAVIS dataset, including \emph{Breakdance}, \emph{Camel}, \emph{Bmx}, \emph{Car-roundabout}, \emph{Car-shadow}, and \emph{Dancejump}.}
    \label{apdx:qualitative}
\end{figure*}

We present additional qualitative results on a broader set of videos, demonstrating the effectiveness of our method in capturing fine-grained details. Specifically, as shown in \cref{apdx:qualitative}, in the \emph{Breakdance} video, our model distinctly reconstructs the lettering on the T-shirt, which is not well-represented by other models. Similarly, in the \emph{Car-roundabout} video, unlike other methods, our model accurately reconstructs the ``P'' sign and the structural shapes. Moreover, in the \emph{Bmx} and \emph{Car-shadow} videos, the details of the wheels are more precisely captured by our method. Notably, in \emph{Car-shadow}, the shadows are faithfully reproduced, showcasing our model's ability to handle subtle visual features. These qualitative results highlight the capability of Gaussian representations to better capture fine-grained details compared to other approaches.

\end{document}